\documentclass{llncs}
\usepackage{mathptmx}
\usepackage{setspace}
\usepackage{textcomp}
\usepackage[english]{babel}
\usepackage[latin1]{inputenc}
\usepackage{verbatim}
\usepackage{fancyvrb}
\usepackage{graphics}
\usepackage{graphicx}
\usepackage{algorithm}
\usepackage{algpseudocode}
\usepackage{url}
\usepackage{cite}
\usepackage{stfloats}
\usepackage{longtable}
\usepackage{multirow}
\usepackage{hhline}
\usepackage{makecell}
\usepackage[table,xcdraw]{xcolor}

\usepackage{hyperref}
\usepackage{color}
\usepackage[caption=false]{subfig}

\newcommand{\keywords}[1]{\par\addvspace\baselineskip
\noindent\keywordname\enspace\ignorespaces#1}
\def\FW{\textsc{Prof-Onto}} 
\def\ONT{\textsc{OASIS}}

\DeclareMathAlphabet\mathbfcal{OMS}{cmsy}{b}{n}
\let\mathcal\undefined \DeclareMathAlphabet{\mathcal}{OMS}{cmsy}{m}{n}
\newcommand{\D}{\mathbf{D}}

\newcommand{\sroiqd}{\mathcal{SROIQ}(\D)}

\begin{document}

%\doublespace
\begin{frontmatter} 
\title{Ontological Smart Contracts in OASIS: Ontology for Agents, Systems, and Integration of Services (Extended Version)}

\author{
{Domenico Cantone, Carmelo Fabio Longo,
		Marianna Nicolosi-Asmundo,\\ Daniele Francesco Santamaria, Corrado Santoro\\ }
\institute{University of Catania\\
		Department of Mathematics and Computer Science\\
		Viale Andrea Doria, 6 - 95125 - Catania, ITALY\\\{domenico.cantone, fabio.longo\}@unict.it,\\ \{nicolosi, santamaria, santoro\}@dmi.unict.it}
}

\maketitle

\begin{abstract}
In this contribution we extend an ontology for modelling agents and their interactions, called \emph{Ontology for Agents, Systems, and Integration of Services} (in short, \ONT{}), with conditionals and \emph{ontological smart contracts} (in short, OSCs). OSCs are ontological representations of smart contracts that allow to establish responsibilities and authorizations among agents and set agreements, whereas conditionals allow one to restrict and limit agent interactions,  define activation mechanisms that trigger agent actions, and define constraints and contract terms on OSCs. Conditionals and OSCs, as defined in \ONT, are applied to extend with ontological capabilities digital public ledgers such as the \emph{blockchain} and  smart contracts implemented on it. We will also sketch the architecture of a framework based on the \ONT{} definition of OSCs that exploits the \emph{Ethereum} platform and the \emph{Interplanetary File System}.    
\end{abstract} 

%\begin{keyword}
%MAS \sep Agent \sep IPFS \sep Smart Contract \sep Ontologies \sep Blockchain \sep Ethereum
%\end{keyword}

\end{frontmatter}

%%
%% intro.tex
%%
%\documentclass[main-mimmo.tex]{subfiles}
%\begin{document}
%\doublespace

\section*{Introduction}
\label{sect:intro}

One of the most important features of a decentralized and publicly shared ledger is the elimination of any third-party intermediaries, since they require clients to put total and unquestioned trust on them. This is particularly true when the ledger is also responsible for managing legal contracts in a digital platform. The \emph{blockchain}~\cite{christidis16} has been introduced to allow parties of a network to interact in a distributed manner without the requirement of trusted entities. The blockchain  is a peer-to-peer public ledger maintained by a distributed network of computational nodes. Blockchain technologies are opening new perspectives in several key aspects such as \emph{Internet of Things} (IoT), healthcare, insurance, energy,  communications,  and robotics~\cite{xu2017taxonomy}, in view of the wide range of benefits they provide. For example, the blockchain guarantees the ownership, transparency, traceability, public availability, continuity, and immutability of digital  assets, in an efficient and trust-less environment where censorship is not achievable. One of the most popular applications of the blockchain is a self-executable contract, also called \emph{smart contract} (SC)~\cite{Szabo97}, a way of representing contracts into lines of immutable program codes which are allowed to be self-run on a public ledger.

Smart contracts on the blockchain are equivalent to \emph{stored procedures} of databases, and hence they have direct access to low level mechanisms. Abstractions of smart contracts may provide several advantages. Among these, we recall that it is easier to publish contracts that integrate and operate with several types of digital and non-digital assets, when higher-level and formal representation of their constraints and agreements are provided. In addition, high-level representations of smart contracts are easy readable by human agents, thus enabling a clear understanding of the agreements and the verification of violations outside the digital borders of the application, for example, in a lawsuit context. Finally, such contracts are platform independent, and hence they can be used and shared by several types of applications and systems.

Semantic web tools and languages aim to reach full machine interoperability, to promote common data formats, and to  exchange protocols on the web, share and reuse data across applications and across enterprise and community boundaries. In the semantic vision of the web,  software agents are enabled to query and manipulate information on  behalf of human agents by means of machine-readable data that carry explicit meaning. Thus, data can be automatically processed and integrated by agents, and can be accessed and modified at a higher level in such a way as to increase coherence and dissemination of information. In addition, with the intervention of semantic reasoners,  implicit information is processed and inferred  as to gain a deeper knowledge of the domain.  Moreover, automated reasoning systems allow one to also verify consistency of data and query it. The definition of a specific domain is widely called \emph{ontology}~\cite{oberle09,hofweber18}. The \emph{Word  Wide  Web Consortium} (W3C) recommends the \emph{Web Ontology Language 2} (OWL 2), a knowledge representation language for web ontologies relying on the Description Logic $\sroiqd\space$\cite{Horrocks2006}.

% \ONT{} is a foundational OWL 2 ontology modelling relevant aspects of multi-agent systems (MAS) such as: a) behaviors of agents, which are conceived in terms of operations that agents are  able to perform; b) templates of behaviors, used to build default agent behaviors;  c)  configurations of agents and their controlled component; d) requests of agents for the execution of actions; e) entrustments of agents with performing actions;  f) execution of operations and their related.

In \cite{woa2019}, we presented the \textit{Ontology for Agents, Systems, and Integration of Services}\footnote{http://www.dmi.unict.it/\texttildelow santamaria/projects/oasis/oasis.php} (in short, \ONT{}), a foundational OWL 2 ontology that defines a request-execute communication protocol for agents, and in particular for \emph{Internet of Agents} (IoA), based on a mutual exchange of ontology fragments that allow a full transparent and high-level interoperability of agents. For example, such ontology fragments allow agents to send and retrieve information from other agents in a transparent way, request other agents the execution of operations without knowing their hardware and software specifications,  acknowledge agents for the execution status of the requested actions, etc. Moreover,  \ONT{} models information about the assignment and the execution of operations, restrictions on requests, connections, exchange of messages among agents, etc.

In this paper, we move towards the definition of a semantic blockchain by extending \ONT{} so as to deal with agreements that can be established by agents an secured on the blockchain without requiring the definition of specific blockchain smart contracts. Such agreements, called \emph{ontological smart contracts} (OSCs),  are established by leveraging conditionals that are also used to abstract and formalize behavior constraints, ways of limiting, bounding, or triggering agent actions. OSCs can be secured  through suitable smart contracts implemented on the blockchain and allowing storing and retrieving of RDF graphs. We will also sketch the architecture of such a system that exploits the new introduced features of \ONT{} and based on the \emph{Interplanetary File System} (IPFS) \cite{ipfs}, the latter allowing blockchain-based applications to use immutable files of large dimensions without excessively increasing the cost of transactions in terms of computation and crypto-currency. Finally, we provide an implementation on Ethereum of the designed architecture.

The paper is organized as follows. Section \ref{sect:related} deals with related works.  Section \ref{sect:arch} is devoted to the description of ontological smart contracts in \ONT{}, whereas in Section \ref{sect:case} we sketch the architecture of a framework based on OSCs exploiting the blockchain and the IPFS. Finally, Section \ref{sec:concl} concludes the paper with some final considerations and hints for future work.

\section{Related work}
\label{sect:related}

Since 2000, several results concerning how agents enter and leave different interaction systems have been presented, by exploiting both \emph{commitment objects} \cite{fornara2004a} and \emph{virtual institutions} \cite{esteva2003}. Only very recently, however, researchers have focused their interest in conjoining the blockchain and ontologies  \cite{CanoCimmino19}. One of the areas of
investigation has been the development of a characterization of blockchain concepts and technologies through ontologies. A theoretical approach looking at the blockchain with an ontological approach has been introduced in \cite{Kruijff2017}, whereas \cite{RSICPD18} proposes a blockchain framework for \emph{semantic web of things} (SWoT) contexts settled as a \emph{Service-Oriented Architecture} (SOA), where nodes can exploit smart contracts for registration, discovery, and selection of annotated services and resources.

Other works aim to represent ontologies within a blockchain context. In \cite{KimL18}, ontologies are used as a common data format for blockchain-based applications, though limited to the description of implementation aspects of the blockchain. However, some of  the architectural choices presented in this paper to effectively implements OSCs through blockchain are inspired from \cite{KimL18}.  

\begin{sloppypar}
The \emph{Blockchain Ontology with Dynamic Extensibility} (BLONDiE)  project \cite{Rojas2017AMP} provides a comprehensive vocabulary that covers the structure of different components (wallets, transactions blocks, accounts, etc.) of blockchain platforms (Bitcoin and Ethereum)  and that can be easily extended to other alternative systems. 
\end{sloppypar}

In \cite{Fill2019ApplyingTC}, the authors discuss the (possible) applications of the blockchain for tracking the provenance of knowledge, for establishing delegation schemes, and for ensuring the existence of patterns in formal conceptualizations using zero-knowledge proofs. 

We proposed in \cite{woa2019} a first version of \ONT{}, a behavior-oriented ontology for representing agents and their interactions. As far as we know, this represents the first attempt of using semantic web technologies for defining a transparent communication protocol among agents that abstracts from their implementation details and configurations, for integrating and uniforming agents, and as a representation system for their behaviors and interactions.
In \cite{woa2019}, we also proposed a prototype version based on two modules of a domotic assistant, called \FW, that  exploits several features of \ONT{} in order to enable communication of IoT devices and users in a domotic environment. The first module of \FW{} is the ontological core, implemented  in Java by exploiting the OWL API \cite{owlapi} and Apache Jena  framework \cite{jena} to manipulate the ontological information and to perform SPARQL queries. %Consistency of \FW{} knowledge bases is checked by means of the HermiT DL reasoner~\cite{hermit}.
The  second  module is  a  proactive  daemon, written in Python, which is used to  engage  devices and to implement the domotic assistant and the interface of \FW{}.
%\end{document}

%\input{prelim.tex}
%%
%% arch.tex
%%
%%
%\documentclass[main-mimmo.tex]{subfiles}
%\begin{document}
%\doublespace

\section{Ontological smart contracts in \ONT{}} \label{sect:arch}

In the last years, digital contracts became very popular in many ambits such as healthcare,  digital security, economics, and business, since they provide traceability, transparency, and irreversibility of transactions. SCs play a prominent role in such contexts. SCs are computer protocols intended to digitally publish, verify, and enforce the negotiation or the performance of self-executing contracts in a distributed and decentralized way, where the agreement between parties is directly written into lines of programming code. Such a transposition of  classical contracts in program codes performs the corresponding actions upon automatically verifying whether the conditions of the contract are satisfied. Basically,  SCs exploit  fragments of programming code of a selected language that automatically  perform the satisfiability check of the contract agreements and verify that the information specified by the clauses stipulated by the involved parts correspond to the reality.  Transparency and traceability of every kind of expected operations is guaranteed by a decentralized public ledger  that can be a suitable platform for holding contracts between suppliers and public administrations, under the eyes of supervisors of upper levels of governance. Every aspect of every-day life  may be aware of public contracts by means of appropriate tools somehow directly connected to the blockchain. In the context of multi-agent systems, for instance, SCs may guarantee agreements and responsibilities among agents, as shown in the case study illustrated in Sections \ref{sect:arch} and \ref{sect:case}.

\smallskip

Before introducing the classes and properties adopted by \ONT{} to represent smart contracts, we first present \textit{conditionals}. %, which are ways used in this context to characterize contract terms. A contract term is a clearly expressed provision giving rise to an obligation and whose breach can lead to a litigation. 
In general, conditionals are used outside the context of digital contracts  to put constraints on the execution of actions and to ensure that some conditions are verified before executing a task.  Conditionals are  OWL sentences that have the fashion of  \emph{Semantic Web Rule Language} (SWRL) rules \cite{swrlbook} describing operations that must be triggered when certain conditions hold and exchanged among agents. In fact, unlike SWRL rules which check the consistency of the knowledge base and infer new sentences, OWL representation of conditionals allows \ONT{}-oriented applications and systems to combine constraints with behaviors, thus extending the expressiveness and providing a higher-level description and representation of agents. Moreover, the introduction of conditionals in \ONT{} is also justified by the fact that the satisfiability  of conditionals does not coincides with the satisfiability of the knowledge base. In fact, violations of conditionals may result in  a litigation among agents or may lead agents to actuate alternative plans, without  affecting in any way the satisfiability of the knowledge base. In the context of smart contracts, conditionals are used to characterize contract terms which are clearly expressed provisions that give rise to an obligation and whose breach can lead to a litigation.

\begin{figure}[H]
	\centering
	\fbox{\includegraphics[scale=0.5]{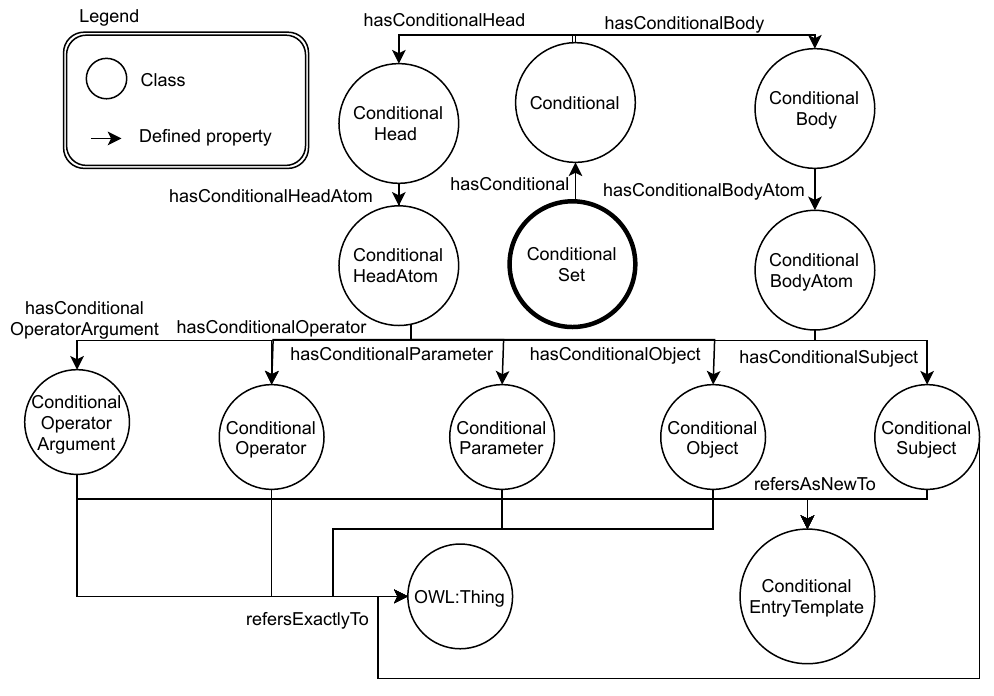}}
	\caption{Ontology schema of conditionals}
	\label{fig:conditional-schema}
\end{figure}

Fig. \ref{fig:conditional-schema} depicts the schema of \ONT{} conditionals. Conditional are constituted by  a consequent (head) and an antecedent (body),  both formed by a conjunctive set of atoms. Atoms in their turn comprise the subject of the conditional, the object, the operator describing the action and, possibly, an operator parameter and  argument.
Conditional atoms are introduced by means of the following classes:

\begin{itemize}
    \item[-] \textit{ConditionalAtom}: represents a conditional atom;
    
    \item[-] \textit{ConditionalHeadAtom}: represents atoms of conditional consequents and is defined as a subclass of the class \textit{ConditionalAtom};
    
    \item[-] \textit{ConditionalBodyAtom}: represents atoms of conditional antecedents and is defined as a subclass of the class \textit{ConditionalAtom};
    
    \item[-] \textit{ConditionalSubject}: represents subjects of atoms of conditional consequents or of conditional antecedents;
    
    \item[-] \textit{ConditionalObject}: represents objects of atoms of conditional consequents or of conditional antecedents;
    
    \item[-] \textit{ConditionalOperator}: represents actions of atoms of conditional consequents or of conditional antecedents;
    
    \item[-] \textit{ConditionalParameter}: represents parameters of actions considered by conditional consequents or conditional antecedents (the class \textit{ConditionalParameter} includes the classes \textit{ConditionalInputParameter} and \textit{ConditionalOutputParameter}, representing the input and output parameter, respectively);
    
    \item[-] \textit{ConditionalOperatorArgument}: defines operator arguments which represent a subordinate characteristic of the conditional operator (for example, \textquotedblleft quality check'' may be represented by the operator \textit{check} with argument \textit{quality});
    
     \item[-] \textit{ConditionalEntryTemplate}: represents templates of the features that the entities involved in the conditional, which are introduced by means of the object-property \textit{refersAsNewTo}, must satisfy.
\end{itemize}

\vspace*{-0.5cm}
\vspace*{-0.5cm}

\noindent Conditionals are introduced in \ONT{} by means of the following classes:

\begin{itemize}
     \item[-] \textit{Conditional}: defines conditionals, constituted by a consequent (head) and an antecedent (body), both consisting in a conjunctive set of atoms;
     
     \item[-] \textit{ConditionalHead}: represents consequents (heads) of conditionals;
     
    \item[-] \textit{ConditionalBody}: represents  antecedents (bodies) of conditionals;
    
     \item[-] \textit{ConditionalSet}: represents  conjunctive sets of conditionals.
\end{itemize}

%Moreover, instead of adding triples to the knowledge base, \ONT{} conditionals are conceived to provide an abstraction of the operations that agents should trigger upon request, when certain conditions hold. 

%, representing the conditional antecedents and the conditional consequents. %Conditional sets are constituted by  disjunctive conditionals, and hence it is sufficient to satisfy one conditional in order to consider satisfied the conditional set. provide one or more atom of conditional consequent and antecedent connecting instances of the class \textit{Conditional} by means of the object-properties \textit{hasConditionalHead} and \textit{hasConditionalBody}, respectively.

Conditional atoms are related to a subject (instance of the class \textit{ConditionalSubject}), an object (instance of the class \textit{ConditionalObject}), and  an operator (instance of the class \textit{ConditionalOperator}),  by means of the object-properties \textit{hasConditionalSubject}, \textit{hasConditionalObject}, and \textit{hasConditionalOperator}, respectively. Possibly, conditional atoms  are related to parameters (instances of the class \textit{ConditionalParamenter}) and to  operator arguments (instances of the class \textit{ConditionalOperatorArgument}) by means of the object-properties \textit{hasConditionalParameter} and \textit{hasConditionalOperatorArgument}, respectively.  Specifically, input and output parameters are introduced through 
the object-properties \textit{hasConditionalInputParameter} and \textit{hasConditionalOutputParameter}, respectively.

In \ONT{} there are two ways to connect conditional entries (i.e., instances of \textit{ConditionalSubject}, \textit{ConditionalObject}, \textit{ConditionalOperator},  \textit{ConditionalParamenter}, and \textit{ConditionalOperatorArgument}) to the related entities, namely  by exploiting the object-property \textit{refersExactlyTo} and by exploiting the object-property \textit{refersAsNewTo}. 

The first way consists in directly indicating the individual involved in the conditional by means of the object-property \textit{refersExactlyTo}, e.g., the address of a digital wallet of a specific person. In the second modality, the entry is unknown, but there is a set of desirable features that the entry must have in order to make the conditional satisfied when it is checked. For example, if a digital asset is sold to anyone who sends the correct amount of money to a digital wallet, the buyer (the conditional entry) is not known until he performs the transaction: completing a transaction is the feature required in order to consider an entity as a buyer. 
In such a case, the object-property \textit{refersAsNewTo} is used to specify an instance of the class \textit{ConditionalEntryTemplate} which endows the set of features that must be satisfied by the conditional entry.

\begin{sloppypar}
Instances of the classes \textit{ConditionalHead} and \textit{ConditionalBody} are related to instances of the classes \textit{ConditionalHeadAtom} and \textit{ConditionalBodyAtom}, representing conditional atoms through the object-properties \textit{hasConditionalHeadAtom} and \textit{hasConditionalBodyAtom}, respectively. The object-properties \textit{hasConditionalHeadAtom} and \textit{hasConditionalBodyAtom} are defined as subproperties of the object-property \textit{hasConditionalAtom}.
\end{sloppypar}

Finally, conditionals are introduced in \ONT{} by way of instances of the class \textit{ConditionalSet}. Such instances are linked through the object-property \textit{hasConditional} to  instances of the class \textit{Conditional} which, in their turn, are linked to instances of the classes  \textit{ConditionalHead} (representing the consequent) and \textit{ConditionalBody} (representing the antecedent) by means of the object-properties \textit{hasConditionalHead} and \textit{hasConditionalBody}, respectively.

%states that a specific individual  In such a case, the agent that verifies the conditional must look for the entity having the IRI specified by \textsf{I}.
%On the contrary, the object-property \textit{refersAsNewTo} is used to state that the connected entity is not currently present in the knowledge base and that it represents a template or a general schema of the features that the entry must satisfy. Additionally, the entry referred to by means of the object-property \textit{refersAsNewTo} is instance of the class \textit{ConditionalEntryTemplate}. The latter instance will be connected to the actual entity by means of the object-property \textit{implementsConditionalEntry}, at the time of contract instantiations.

\begin{figure}[H]
	\centering
	\fbox{\includegraphics[scale=0.55]{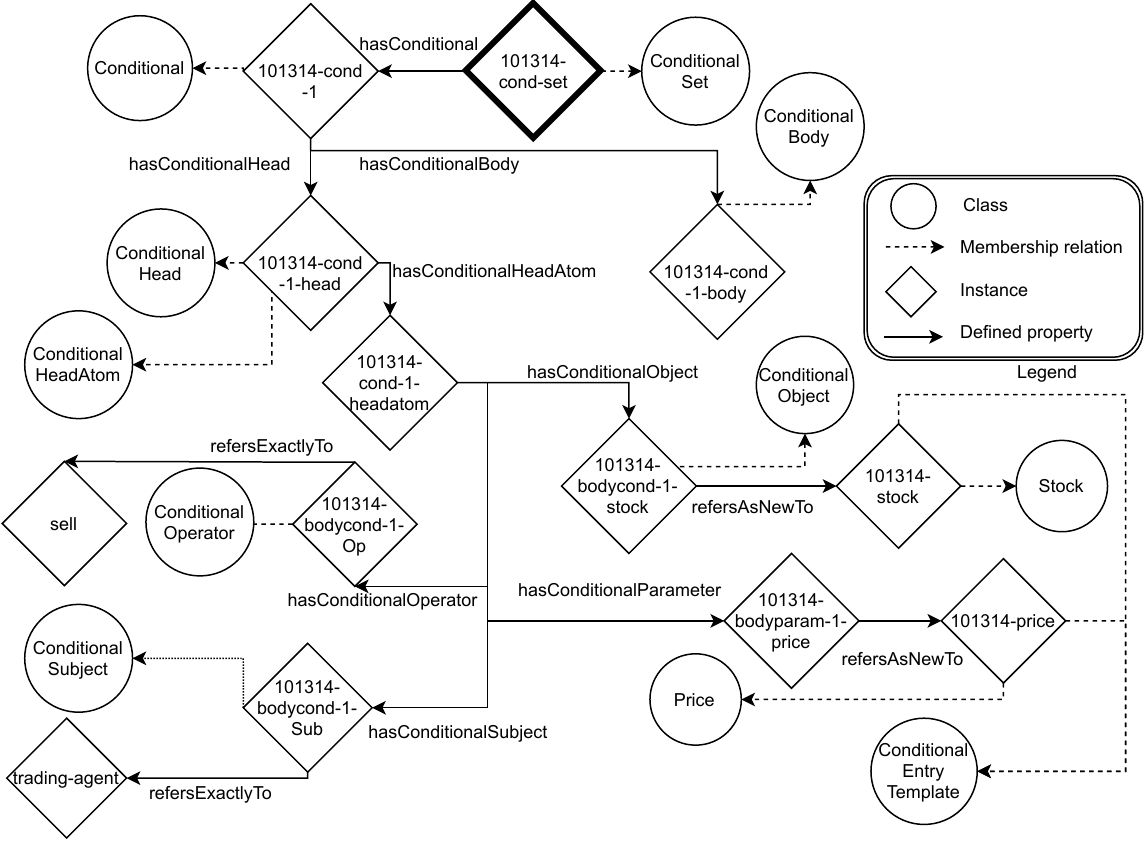}}
	\caption{Example of \ONT{} conditional}
	\label{fig:conditional-example-unique}
\end{figure}

\begin{sloppypar}
In Fig. \ref{fig:conditional-example-unique},  we show an example of conditional used by a trading agent (\textit{trading-agent} to sell (\textit{sell}) stocks (\textit{01314-stock}) on behalf of clients at a certain price (\textit{101314-price}). Specifically, the figure illustrates the main structure of the conditional set, which comprises one conditional admitting, in its turn, one conditional head and one  conditional body. The head consists of a single atom head modelling the selling action,  whereas the body is empty.  %models the dropping of the price. For space reason, Fig. \ref{fig:conditional-example-unique} reports only the  atom head of the conditional, which is the most interesting part.
\end{sloppypar}

\begin{figure}[H]
	\centering
	\fbox{\includegraphics[scale=0.6]{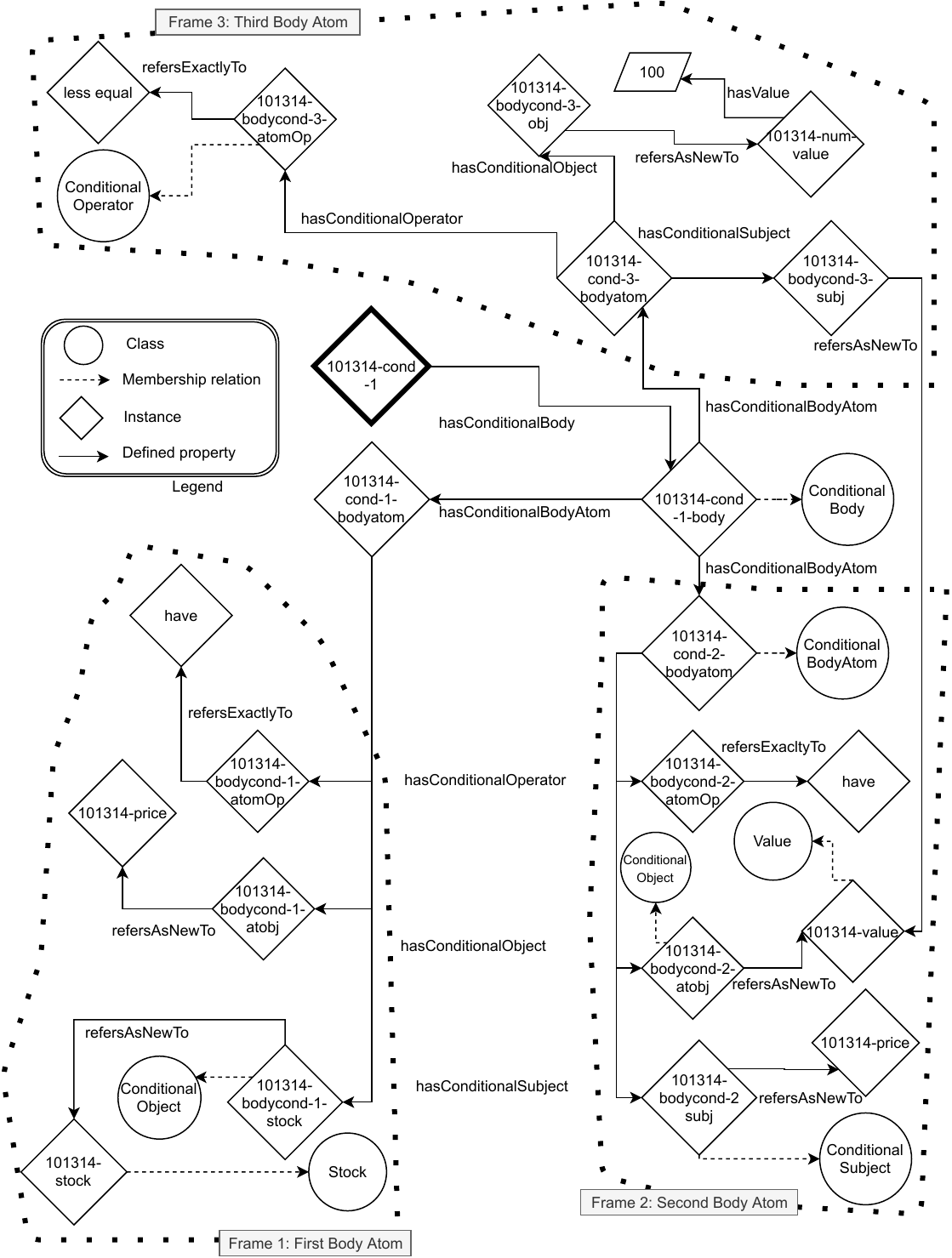}}
	\caption{An extension of the conditional of Fig. \ref{fig:conditional-example-unique}.}
	\label{fig:conditional-example-3}
\end{figure}

The case study depicted in Fig. \ref{fig:conditional-example-unique} can be also extended to design more complex scenarios. For example, the conditional presented in Fig. \ref{fig:conditional-example-unique} can be extend to guarantee that the trading agent sells the stocks when their price drops under $100$ dollars. In such a case, we introduce a new body atom (\textit{101314-cond-1-body}) as depicted in Fig.\ref{fig:conditional-example-3}.

In Fig. \ref{fig:conditional-example-3}, the three body atoms of the conditional are pointed out by three frames. In Frame 1, the price of the stock defined in the head atom is checked (individual \textit{101314-price}). The individual representing the stock and the individual representing the price are the same as in the head atom. A conditional operator linking to the task operator \textit{have} is introduced to state that the stock has a specific price. 

In Frame 2, the value of the price is checked. Specifically, we introduce a new individual, namely, \textit{101314-value}, to represent the current value of the price. The task operator \textit{have} is used again to state that the price has the indicated specific value.

Finally, in Frame 3, we ensure that the  value \textit{101314-value} is related with the numeric value $100$ by directly connecting it with the integer by means of the object-property \textit{value}. Exploiting the task operator \textit{less equal}, connected to the conditional operator through the object-property \textit{refersExactlyTo}, we ensure that the effective numeric value associated to the current value of the price is less equal that $100$.

\vspace*{-0.5cm}
\vspace*{-0.5cm}

\clearpage

\begin{figure}[H]
	\centering
	\fbox{\includegraphics[scale=0.7]{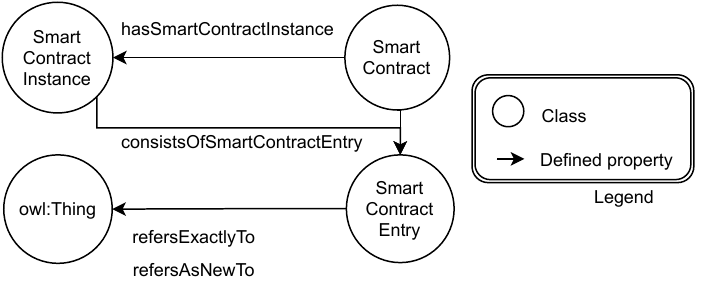}}
	\caption{Ontology schema of smart contract}
	\label{fig:sc-schema}
\end{figure}

\begin{sloppypar}
Smart contracts benefit from the abstraction layer provided by suitable ontological models since they are defined at a higher level, leaving the implementation details to the underneath layer constituted by blockchain-based distributed computing platforms such as \emph{Ethereum}.\footnote{https://www.ethereum.org} With this in mind, conditionals are exploited to define ontological smart contracts in \ONT{} according to the schema illustrated in Fig. \ref{fig:sc-schema}.
\end{sloppypar}

%\begin{sloppypar}
Anytime two parties agree to take action, in order to make an exchange or payment for something of value, a contract is created (instantiated). We call such a contract actuation a \emph{contract instance}. For instance, in the case of a trading agent, the contract specifies a template of the general agreements that trader and clients shall respect, whereas the contract instance is the application of such a template, specifying how the two parties join the contract. For example, a trader  and a client may instantiate a general brokerage contract, where the trader is responsible for selling stock on behalf of clients.
%car lease contract, the contract specifies a template of general agreements that lessee and lessor shall respect, whereas the contract instance is the application of such a template between a specific lessee and lessor, specifying how the two parties join the contract.  
In \ONT{}, smart contracts are introduced by the class \textit{SmartContract}, whereas contract instances are modelled by means of the  class \textit{SmartContractInstance}. Smart contracts and smart contract instances are connected through the object-property \textit{consistsOfSmartContractInstance}.
%Instances of the class \textit{SmartContract} (resp., \textit{SmartContractInstance}) are linked by means of the object-property \textit{hasSmartContractType} to instances of the class \textit{SmartContractType}, representing the type of the smart contract (resp., of the smart contract instance).

Smart contracts and smart contract instances provide a set of entries mapped by instances of the class \textit{SmartContractEntry}, the latter containing the classes \textit{SmartContractEntryParticipant} (including individuals referring to the participants involved in the smart contract) and \textit{SmartContractEntryValue} (including individuals referring to values involved in the smart contract). The object-property \textit{consistsOfSmartContractEntry} links smart contracts and  smart contract instances  to the corresponding entries.  Entries of the contract instances are connected to the corresponding contract entries by means of the object-property \textit{refersExactlyTo}.

 % the latter connected to the entity joining the agreement by means of the object-property \textit{refersExactlyTo}.
%\end{sloppypar}
%We always refer to  smart contract instances as \emph{smart contract instances} to avoid confusion with the  term \emph{class instance} adopted in OWL.
 % The object-properties  \textit{refersExactlyTo} and  \textit{refersAsNewTo} are used to connect smart contract entries to the referred entity and to the template entity, respectively, in the same way as in the case of conditionals.
%In \ONT{}, smart contract instances are introduced by means of the set of OWL triples that are in some way related to the representation of the contract and whose presence is justified only in light of the contract creation. For all the triples modelling the contract instance, we consider only the root as the representative of the instance (e.g., the lessee activity). 

\begin{figure}[H]
	\centering
	\fbox{\includegraphics[scale=0.6]{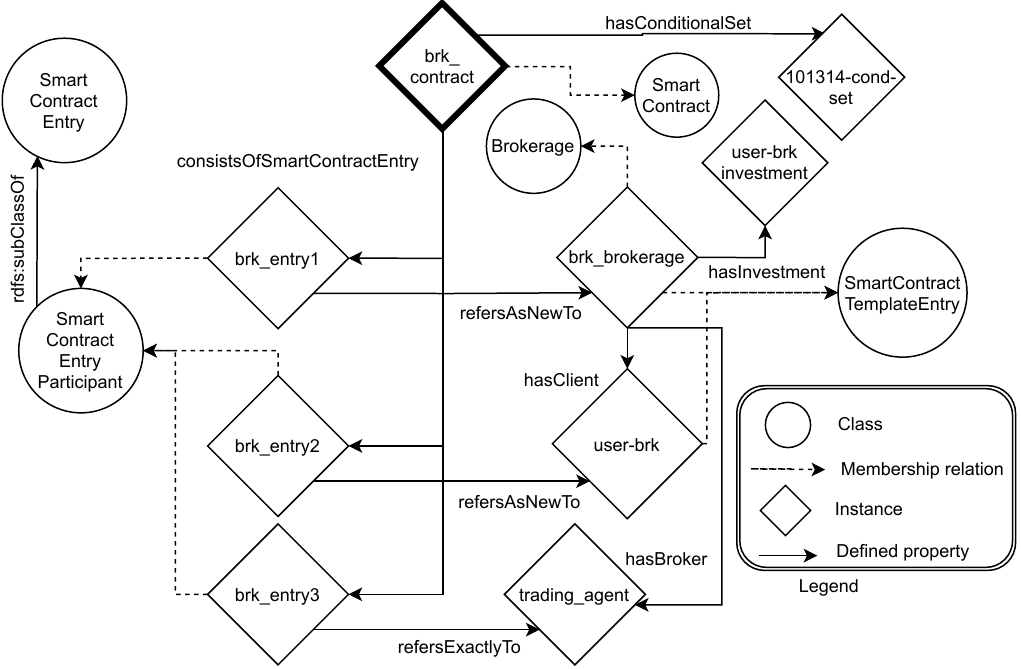}}
	\caption{Example of a smart contract in \ONT}
	\label{fig:sc-example}
\end{figure}

\begin{figure}[H]
	\centering
	\fbox{\includegraphics[scale=0.6]{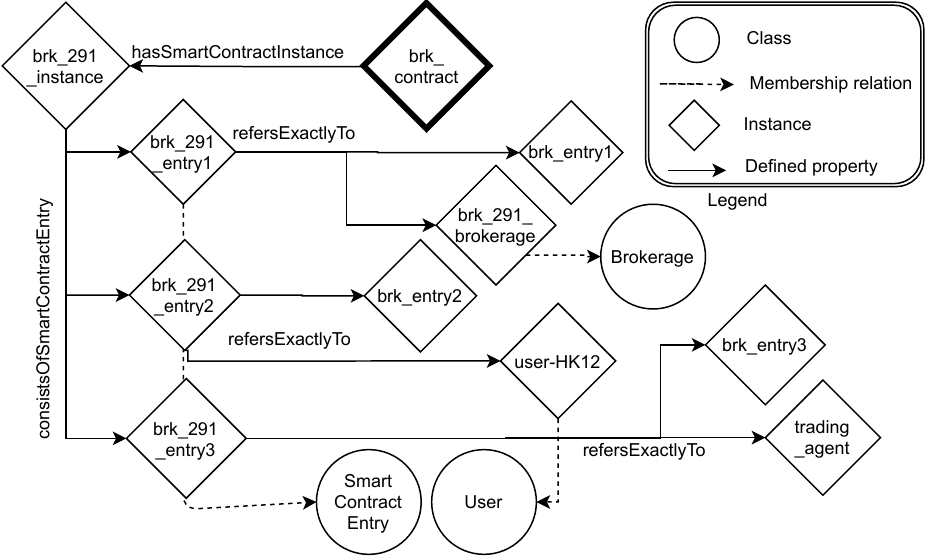}}
	\caption{Example of a smart contract instance in \ONT}
	\label{fig:sc-example-inst}
\end{figure}

In Figs. \ref{fig:sc-example} and \ref{fig:sc-example-inst}, we show a trading contract (stipulated between a trading agent and a potential client introduced by means of a template) and one of its instances.

%In the considered case, the user entrusts a certain investment capital to a trading agent, which guarantees that every stock brought is sold if its price drops under $100$ dollars, as depicted by the conditional in Fig. \ref{fig:conditional-example-1}, \ref{fig:conditional-example-2}, and \ref{fig:conditional-example-3}. 
\begin{sloppypar}
The contract, formed by three entries, is modelled by the instance of the class \textit{SmartContract} called \textit{brk\_contract} (the element in bold in Fig. \ref{fig:sc-example}). The contract involves two participants, the trading agent and a template of a general client, respectively represented by the entities \textit{trading\_agent} and \textit{user-brk}, and a value representing a template of the brokerage activity (\textit{brk\_brokerage}).  Then, the three smart  contract entries are connected to: the template of a brokerage activity, the template of a client (by means of \textit{refersAsNewTo}), and the trading agent (by means of \textit{refersExactlyTo}), respectively. The brokerage activity template also specifies that an investment must be involved in the contract. 
\end{sloppypar}

The contract instance in Fig. \ref{fig:sc-example-inst} introduces three entries by means of the object-property \textit{consistsOfSmartContractEntry}. These, in their turn, are connected to the corresponding individual (the user \textit{user-HK12}, the actual brokerage \textit{brk\_291\_brokerage}, and the trading agent) by means of the object-property \textit{refersExactlyTo}. Moreover, the three entries are related to the corresponding entries of the contract in Fig. \ref{fig:sc-example} by means of the object-property \textit{refersExactlyTo}, in order to specify how contract terms have been defined by the current contract instance. 

Conditionals  are used to characterize constraints among the entries of a smart contract and to establish agreements. According to the will of the user that desires to entrust the trader of selling its stocks, the conditional that models such a situation is added to the contract by connecting the instance \textit{brk\_contract} to the conditional set \textit{101314-cond-set} by means of the object-property \textit{hasConditionalSet} (see Fig. \ref{fig:conditional-example-unique}). Any selling activity concerning the brokerage contract may be easily verified through a SPARQL query that checks whether the stock has been sold (and, additionally, the selling conditions). The results of the query validates (or invalidates) the contract between the two parties. Ontological smart contracts model agreements among parties whereas SPARQL query validate them. What the ontology does not guarantee is that the parties have actually agreed to the clauses of the contract, its traceability, non-repudiation, and so on, features that a blockchain framework, instead, may ensure.

\section{Architectural design of a OSC-oriented application} \label{sect:case}

As stated above, ontological smart contracts may enjoy from a decentralized ledger such as the blockchain. However, it is prohibitively expensive to store a lot of data on it. For instance, at the time of this writing, about 50 US dollars are required  to store  the 38 pages of the PDF version of the Ethereum yellow paper, which weights about 520Kb. In fact, according to the paper itself,  approximately 20,0000 gas are required for storing 256 bit/8 bytes (1 word), namely 20 Gwei for a unit of gas (1 Gwei equals 0.000000001 Ethers). In order to reduce the cost of transactions and the time required to compute them, data-oriented applications need to rely on a decentralized server to store information.  The \emph{Interplanetary File System} (IPFS) is one of the preferred solutions. Basically, IPFS allows one to store a large amount of files, whose permanent IPFS links (CID) can be included into blockchain transactions, in such a way as to  put a timestamp on the data and secure it  without directly including files in the chain itself. It is sufficient to upload documents on IPFS and then to store the IPFS CID on the Ethereum blockchain. The CID is a hash obtained from the file which, hence, cannot be modified.  However, IPFS provides the \emph{InterPlanetary Naming Service} that uses the peer ID to point to a specific hash. Such a hash can change whereas the peer ID cannot. It turns out that applications can gain access to mutable content in IPFS  without  knowing the new hash beforehand.

\vspace*{-0.5cm}
\begin{figure}[H]
	\centering
	\fbox{\includegraphics[scale=0.55]{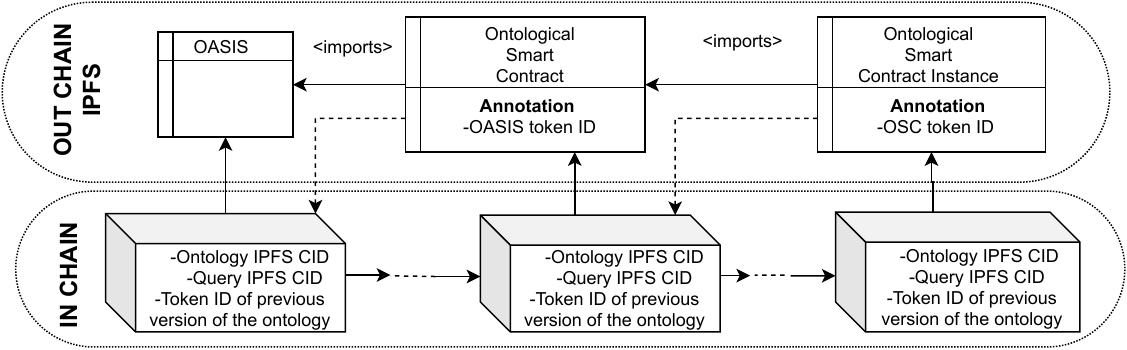}}
	\caption{Sketch of the architecture of an \ONT{} OSC-based application}
	\label{fig:arch-sketch}
\end{figure}
\vspace*{-0.5cm}

\begin{sloppypar}
Fig. \ref{fig:arch-sketch} illustrates a typical application that exploits \ONT{} ontological smart contracts. Ontologies are secured on the blockchain by means of transactions containing only three states: the IPFS CID of the ontology, the IPFS CID of the query required to validate the ontology, and the address of the transaction that secured the previous version of both the ontology and  the corresponding query. As a preliminary step, the ontology \ONT{} and all the ontologies required by the OSC are published on IPFS and secured through a blockchain transaction.\footnote{The smart contract is compliant with the non-fungible token standard ERC721 and is available in the Ethereum main network at the address 0x36194ab80f7649572cab9ec524950df32f638b08. A Java API to publish and retrieve OSC is available at \url{https://github.com/dfsantamaria/CLARA}.} From then on, a second transaction suffices for deploying the OSC. Such a transaction secures the ontology representing the OSC, the query which validates the OSC itself, and the address of the transaction that secured, possibly, a previous version of the OSC and of the query. Analogously, any instance of the smart contract and any related ontology are secured by an additional transaction.
%In analogous way, the OWL serialization of the instances of the ontology is published on IPFS and secured on blockchain
\end{sloppypar}
%For example,  \ONT{} ontology can be accessed by visiting via browser the address \url{https://ipfs.io/ipfs/QmPZ67hyeJPSbegeSitWAQWqqDSUZDPpA5JJ2fJxKHqjA1}.

The IPFS CID of the ontology allows one to access the OSC and to ensure that the OSC and all the imported ontologies have not been modified after their publication, thus guaranteeing that all the ontologies exploited by the contract are exactly the ones on which an agreement has been reached. An additional source of guarantee is the SPARQL query that checks whether the ontological contract is voided by the instance under consideration. As in the case of the ontologies, the SPARQL query is published on IPFS and secured on the blockchain, in the same transaction as the OSC that the query checks.
By accessing the secured OSC and the query which validates it, the participants have all the means to validate or invalidate the contract,  with all the guarantee of the blockchain system and the versatility of ontologies. Moreover, in our architecture, ontologies may exploit suitable OWL annotation  axioms, or alternatively the BLONDiE ontology, to refer to blockchain transactions that secured the imported ontologies.
Finally, the ontological approach and the architecture model introduced in this paper may be also adopted in non-Turing-complete blockchains such as Bitcoin, since the effort required to the blockchain is limited to store at most two states pointing to resources located out of the blockchain, plus one state storing the token ID of the previous version of the ontology.

%\end{document}

%A drawback of such approach may be represented by the OWL serializations. In fact, ones the ontology has been stored on IPFS, the applications that desire to use the related blockchain node must be able to parse the serialization adopted (and we say that the architecture is serialization legacy), since files in IPFS are immutable. Therefore, IPFS provides the \emph{InterPlanetary Naming Service} that uses the peer ID to point to a particular hash. Such hash can change whereas the peer ID doesn't. It turns out that applications can access to mutable content in IPFS  without needing to know the new hash before hand.

%\documentclass[main-mimmo.tex]{subfiles}
%\begin{document}
%\doublespace

\section{Conclusions and Future Work}
\label{sec:concl}
%%
%In this paper we presented \FW{}, a prototype framework that integrates users and IoT devices within domotic environments. \FW{} is based on a novel ontology called \ONT{}, modelling device behaviors, user requests, and their executions. \FW{} acts in two phases. The first phase consists in connecting devices to the domotic assistant. Devices share their behaviors by means of  knowledge bases of \ONT, which are automatically integrated with \FW{} and whose consistency is checked by the HermiT DL reasoner. In the second phase, users send their requests to a BDI rule-based system, called \PROFETA, that maps the transcriptions of user requests in \ONT{} knowledge bases. In the last phase, \FW{}  automatically selects devices compatible with user requests by means of SPARQL queries specifically constructed. Then, the resulting action is sent to the selected device in order to perform the required action. In this phase, a suitable knowledge base of \ONT{} is used as a communication and information exchange system between the assistant and the selected device.

We presented an extension of the \emph{Ontology for Agents, Systems, and Integration of Services} (in short, \ONT{}), modelling ontological conditionals and smart contracts. Conditionals, classically applied to set restrictions on agent actions or to activate them when suitable conditions hold, are also used to define contract terms. Contract terms are applied, in their turn, to define ontological smart contracts, which establish responsibilities and authorizations among agents. Conditionals and smart contracts defined by \ONT{} are exploited to add an ontological level to the blockchain and smart contracts based on it. We also sketched the architecture of a system leveraging the blockchain and the Interplanetary File System (IPFS) to store and retrieve \ONT{} ontological smart contracts, and we implemented it through an Ethereum smart contract.

In order to extend the integration level of \ONT{} with blockchains, we plan to integrate and extend BLONDiE, also by considering the ontology as a meta-model extension of \ONT{}. We also plan  to  study how \ONT{} can be exploited by \emph{OntologyBeanGenerator 5.0} \cite{Briola2018} inside the JADE framework \cite{bellifemine2007}  to generate code for agents and artifacts and how it can be exploited by \emph{CArtAgO} \cite{Ricci2009}, a framework for building shared computational worlds.
%We intend to integrate \ONT{}  with  the ontology for IoT services defined in \cite{wei} and the \emph{Sensor Ontology} in \cite{sensorOnto}. 
We shall extend the set of actions and parameters provided by \ONT{} with the synset introduced by WordNet \cite{wordnet}, in order to make the whole infrastructure multi-language- and meaning-oriented.

Finally, we intend to define a set-theoretic representation of \ONT{} in the flavour of \cite{CanLonNicSanRR2015}. %However, %since  \ONT{} contains existential restrictions, 
\bibliographystyle{ieeetr}
\bibliography{paper}

\end{document}